\documentclass{article} 
\usepackage{url}

\usepackage{bbm}
\usepackage{graphicx}
\usepackage{enumerate}
\usepackage{subcaption}
\usepackage{titlesec}
\usepackage{amsmath,amsfonts,amssymb}
\usepackage{bm}
\usepackage{array}
\usepackage{authblk}
\usepackage{hyperref}

\def\etal{{\textit{et~al.~}}}

\title{Least Squares Generative Adversarial Networks}
\date{\vspace{-5ex}}

\author[1]{Xudong Mao\thanks{xudonmao@gmail.com}}
\author[1]{Qing Li\thanks{itqli@cityu.edu.hk}}
\author[2]{Haoran Xie\thanks{hrxie2@gmail.com}}
\author[3]{Raymond Y.K. Lau\thanks{raylau@cityu.edu.hk}}
\author[4]{Zhen Wang\thanks{zhenwang0@gmail.com}}
\author[5]{Stephen Paul Smolley\thanks{steve@codehatch.com}}

\affil[1]{Department of Computer Science, City University of Hong Kong}
\affil[2]{Department of Mathematics and Information Technology, The Education University of Hong Kong}
\affil[3]{Department of Information Systems, City University of Hong Kong}
\affil[4]{Center for OPTical IMagery Analysis and Learning (OPTIMAL), Northwestern Polytechnical University}
\affil[5]{CodeHatch Corp.}
\begin{document}
\maketitle

\begin{abstract}
Unsupervised learning with generative adversarial networks (GANs) has proven hugely successful. Regular GANs hypothesize the discriminator as a classifier with the sigmoid cross entropy loss function. However, we found that this loss function may lead to the vanishing gradients problem during the learning process. To overcome such a problem, we propose in this paper the Least Squares Generative Adversarial Networks (LSGANs) which adopt the least squares loss function for the discriminator. We show that minimizing the objective function of LSGAN yields minimizing the Pearson $\chi^2$ divergence. There are two benefits of LSGANs over regular GANs. First, LSGANs are able to generate higher quality images than regular GANs. Second, LSGANs perform more stable during the learning process. We evaluate LSGANs on five scene datasets and the experimental results show that the images generated by LSGANs are of better quality than the ones generated by regular GANs. We also conduct two comparison experiments between LSGANs and regular GANs to illustrate the stability of LSGANs.


\end{abstract}

\section{Introduction}
\label{sec:introduction}
Deep learning has launched a profound reformation and even been applied to many real-world tasks, such as image classification ~\cite{He2015}, object detection ~\cite{Ren2015} and segmentation ~\cite{Long2014}. These tasks obviously fall into the scope of supervised learning, which means that a lot of labeled data are provided for the learning processes. Compared with supervised learning, however, unsupervised learning tasks, such as generative models, obtain limited impact from deep learning. Although some deep generative models, e.g. RBM ~\cite{Hinton2006}, DBM ~\cite{Salakhutdinov2009} and VAE ~\cite{Kingma2013}, have been proposed, these models face the difficulty of intractable functions or the difficulty of intractable inference, which in turn restricts the effectiveness of these models. 

Recently, Generative adversarial networks (GANs)~\cite{Goodfellow2014} have demonstrated impressive performance for unsupervised learning tasks. Unlike other deep generative models which usually adopt approximation methods for intractable functions or inference, GANs do not require any approximation and can be trained end-to-end through the differentiable networks.  The basic idea of GANs is to simultaneously train a discriminator and a generator: the discriminator aims to distinguish between real samples and generated samples; while the generator tries to generate fake samples as real as possible, making the discriminator believe that the fake samples are from real data. So far, plenty of works have shown that GANs can play a significant role in various tasks, such as image generation ~\cite{Nguyen2016}, image super-resolution ~\cite{Ledig2016}, and semi-supervised learning ~\cite{Salimans2016}.

\begin{figure*}[t]
\centering
\begin{tabular}{ccc}

 \includegraphics[width=0.3\textwidth]{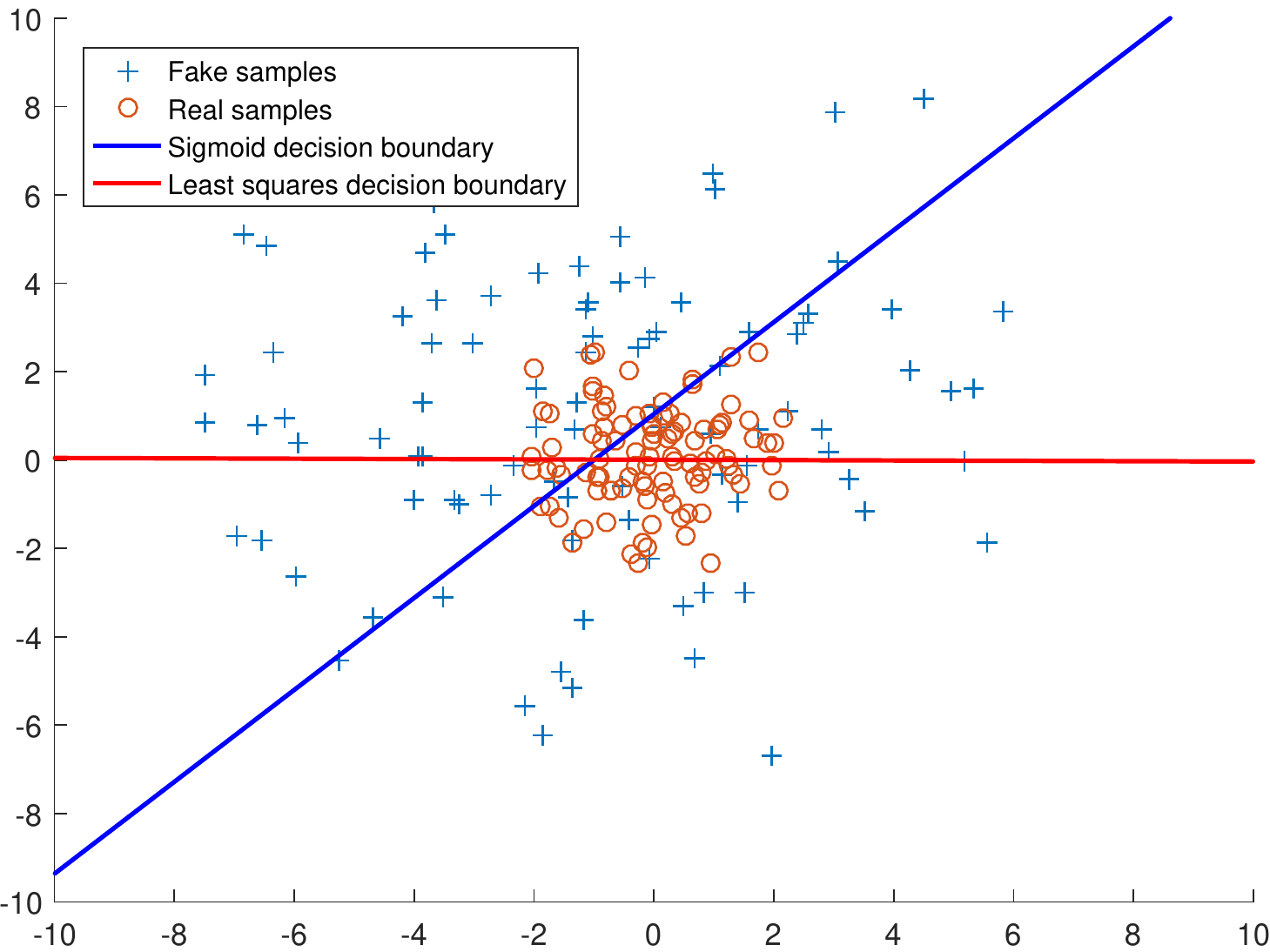}
 &
 \includegraphics[width=0.3\textwidth]{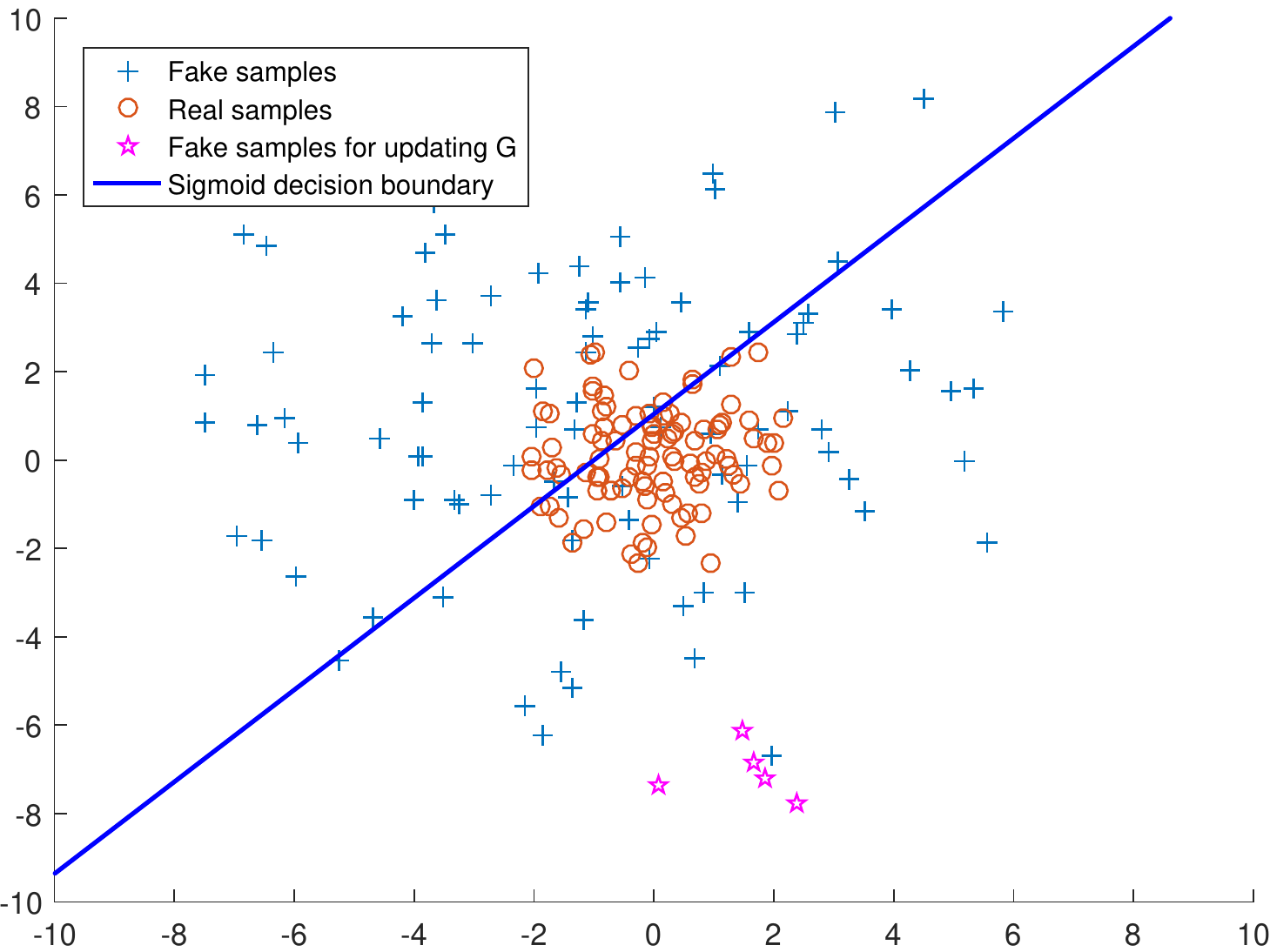}
  &
 \includegraphics[width=0.3\textwidth]{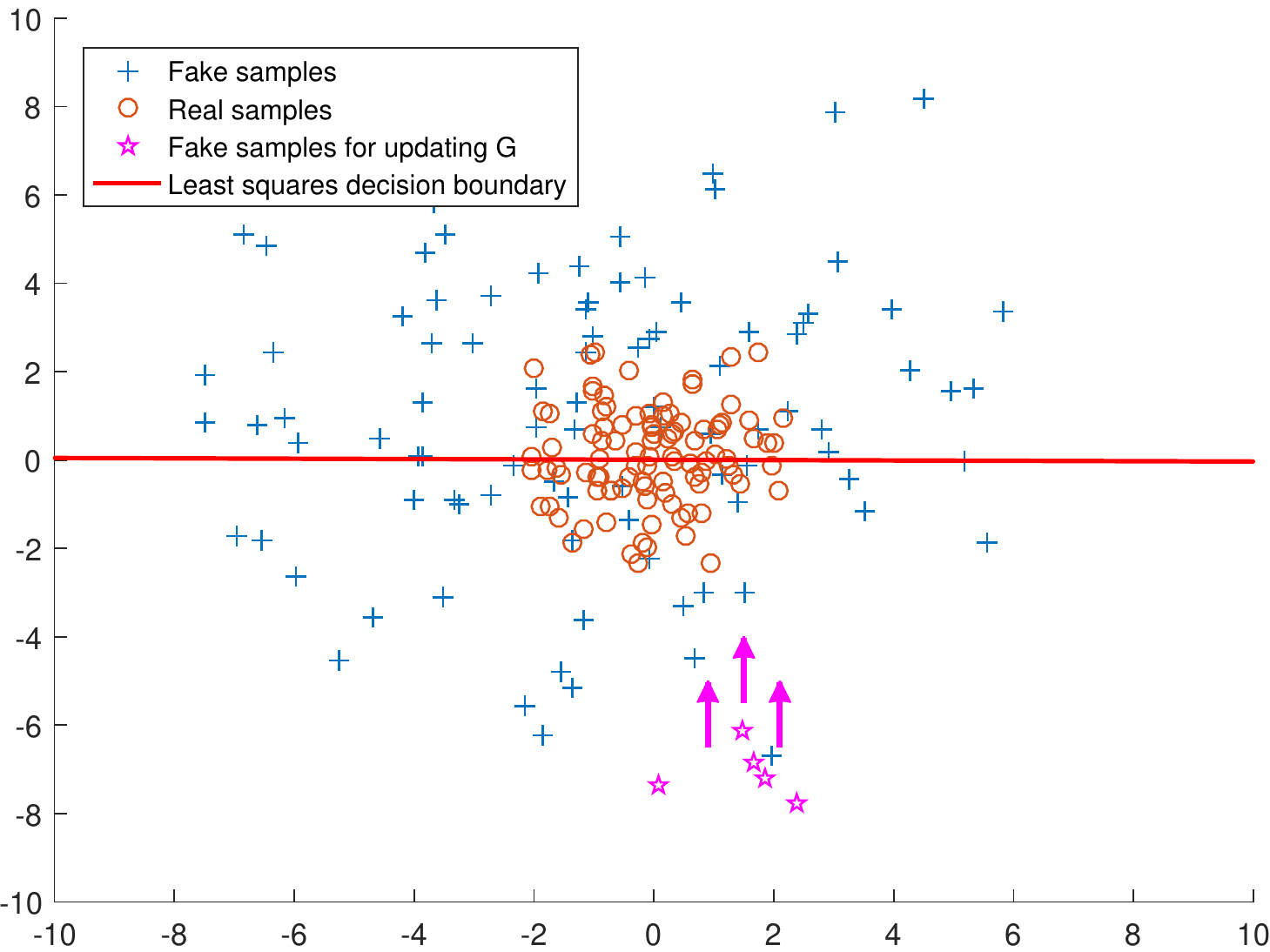}
\\
(a)
&
(b)
&
(c)
\end{tabular}
\caption{
Illustration of different behaviors of two loss functions. (a): Decision boundaries of two loss functions. Note that the decision boundary should go across the real data distribution for a successful GANs learning. Otherwise, the learning process is saturated. (b): Decision boundary of the sigmoid cross entropy loss function. It gets very small errors for the fake samples (in magenta) for updateing G as they are on the correct side of the decision boundary. (c): Decision boundary of the least squares loss function. It penalize the fake samples (in magenta), and as a result, it forces the generator to generate samples toward decision boundary.
}
\label{fig:boundary}
\end{figure*}

In spite of the great progress for GANs in image generation, the quality of generated images by GANs is still limited for some realistic tasks. Regular GANs adopt the sigmoid cross entropy loss function for the discriminator~\cite{Goodfellow2014}. We argue that this loss function, however, will lead to the problem of vanishing gradients when updating the generator using the fake samples that are on the correct side of the decision boundary, but are still far from the real data. As Figure \ref{fig:boundary}(b) shows, when we use the fake samples (in magenta) to update the generator by making the discriminator believe they are from real data, it will cause almost no error because they are on the correct side, i.e., the real data side, of the decision boundary. However, these samples are still far from the real data and we want to pull them close to the real data. Based on this observation, we propose the Least Squares Generative Adversarial Networks (LSGANs) which adopt the least squares loss function for the discriminator. The idea is simple yet powerful: the least squares loss function is able to move the fake samples toward the decision boundary, because the least squares loss function penalizes samples that lie in a long way on the correct side of the decision boundary. As Figure \ref{fig:boundary}(c) shows, the least squares loss function will penalize the fake samples (in magenta) and pull them toward the decision boundary even though they are correctly classified. Based on this property, LSGANs are able to generate samples that are closer to real data. 

Another benefit of LSGANs is the improved stability of learning process. Generally speaking, training GANs is a difficult issue in practice because of the instability of GANs learning~\cite{Radford2015}. Recently, several papers have pointed out that the instability of GANs learning is partially caused by the objective function~\cite{Arjovsky2017, Metz2016,Qi2016}. Specifically, minimizing the objective function of regular GAN suffers from vanishing gradients, which makes it hard to update the generator. LSGANs can relieve this problem because LSGANs penalize samples based on their distances to the decision boundary, which generates more gradients to update the generator. Recently, Arjovsky \etal~\cite{Arjovsky2017} have proposed a method to compare the stability of GANs learning by excluding batch normalization~\cite{Ioffe2015}. Following this method for comparing the stability, we find that LSGANs are also able to converge to a relatively good state without batch normalization.


Our contributions in this paper can be summarized as follows:
\begin{itemize}
\item We propose LSGANs which adopt least squares loss function for the discriminator. We show that minimizing the objective function of LSGAN yields minimizing the Pearson $\chi^2$ divergence. The experimental results demonstrate that LSGANs can generate more realistic images than regular GANs. Numerous comparison experiments are also conducted to prove the stability of LSGANs.
\item Two network architectures of LSGANs are designed. The first one is for image generation with $112 \times 112$ resolution, which is evaluated on various kinds of scene datasets. The experimental results show that this architecture of LSGAN can generate higher quality images than the current state-of-the-art method. The second one is for tasks with a lot of classes. We evaluate it on a handwritten Chinese character dataset with $3470$ classes, and the proposed model is able to generate readable characters.
\end{itemize}

The rest of this paper is organized as follows. Section \ref{sec:related} briefly reviews related work of generative adversarial networks. The proposed method is introduced in Section \ref{sec:method}, and experimental results are presented in Section \ref{sec:experiments}. Finally, we conclude the paper in Section \ref{sec:conclusion}.

\section{Related Work}
\label{sec:related}
Deep generative models attempt to capture the probability distributions over the given data. Restricted Boltzmann Machines (RBMs) are the basis of many other deep generative models, and they have been used to model the distributions of images~\cite{Taylor2010} and documents~\cite{Hinton2009}. Deep Belief Networks (DBNs)~\cite{Hinton2006_DBN} and Deep Boltzmann Machines (DBMs)~\cite{Salakhutdinov2009} are extended from the RBMs. The most successful application of DBNs is for image classification~\cite{Hinton2006_DBN}, where DBNs are used to extract feature representations. However, RBMs, DBNs and DBMs all have the difficulties of intractable partition functions or intractable posterior distributions, which thus use the approximation methods to learn the models. Another important deep generative model is Variational Autoencoders (VAE)~\cite{Kingma2013}, a directed model, which can be trained with gradient-based optimization methods. But VAEs are trained by maximizing the variational lower bound, which may lead to the blurry problem of generated images. 

Recently, Generative Adversarial Networks (GANs) have been proposed by Goodfellow \etal ~\cite{Goodfellow2014}, who explained the theory of GANs learning based on a game theoretic scenario. Compared with the above models, training GANs does not require any approximation method. Like VAEs, GANs also can be trained through differentiable networks. Showing the powerful capability for unsupervised tasks, GANs have been applied to many specific tasks, like image generation~\cite{Chen2016}, image super-resolution~\cite{Ledig2016},  text to image synthesis~\cite{Reed2016} and image to image translation~\cite{Isola2016}. By combining the traditional content loss and the adversarial loss, super-resolution generative adversarial networks~\cite{Ledig2016} achieve state-of-the-art performance for the task of image super-resolution. Reed \etal ~\cite{Reed2016} proposed a model to synthesize images given text descriptions based on the conditional GANs~\cite{Mirza2014}. Isola \etal~\cite{Isola2016} also used the conditional GANs to transfer images from one representation to another. In addition to unsupervised learning tasks, GANs also show potential for semi-supervised learning tasks. Salimans \etal ~\cite{Salimans2016} proposed a GAN-based framework for semi-supervised learning, in which the discriminator not only outputs the probability that an input image is from real data but also outputs the probabilities of belonging to each class.

Despite the great successes GANs have achieved, improving the quality of generated images is still a challenge. A lot of works have been proposed to improve the quality of images for GANs. Radford \etal~\cite{Radford2015} first introduced convolutional layers to GANs architecture, and proposed a network architecture called deep convolutional generative adversarial networks (DCGANs). Denton \etal~\cite{Denton2015} proposed another framework called Laplacian pyramid of generative adversarial networks (LAPGANs). They constructed a Laplacian pyramid to generate high-resolution images starting from low-resolution images. Further, Salimans \etal ~\cite{Salimans2016} proposed a technique called feature matching to get better convergence. The idea is to make the generated samples match the statistics of the real data by minimizing the mean square error on an intermediate layer of the discriminator.

Another critical issue for GANs is the stability of learning process. Many works have been proposed to address this problem by analyzing the objective functions of GANs ~\cite{Arjovsky2017,Che2016,Metz2016,Nowozin2016,Qi2016}. Viewing the discriminator as an energy function, ~\cite{Zhao2016} used an auto-encoder architecture to improve the stability of GANs learning. To make the generator and the discriminator be more balanced, Metz \etal~\cite{Metz2016} created a unrolled objective function to enhance the generator. Che \etal~\cite{Che2016} incorporated a reconstruction module and use the distance between real samples and reconstructed samples as a regularizer to get more stable gradients. Nowozin \etal~\cite{Nowozin2016} pointed out that the objective of the original GAN~\cite{Goodfellow2014} which is related to Jensen-Shannon divergence is a special case of divergence estimation, and generalized it to arbitrary f-divergences~\cite{Nguyen2010}. Another generalization work is presented in literature~\cite{Mohamed2016}. They framed GANs as algorithms for learning in implicit generative models and presented several kinds of probability-based learning losses. Arjovsky \etal~\cite{Arjovsky2017} analyzed the properties of four different divergences or distances over two distributions and concluded that Wasserstein distance is nicer than Jensen-Shannon divergence. Qi ~\cite{Qi2016} proposed the Loss-Sensitive GAN whose loss function is based on the assumption that real samples should have smaller losses than fake samples and proved that this loss function has non-vanishing gradient almost everywhere.

\section{Method}
\label{sec:method}
In this section, we first review the formulation of GANs briefly. Next, we present the LSGANs along with their benefits in Section \ref{sec:lsgan}. Finally, two model architectures of LSGANs are introduced in \ref{sec:arch}.

\begin{figure}[t]
\centering
\begin{tabular}{cc}
 \includegraphics[width=1.5in]{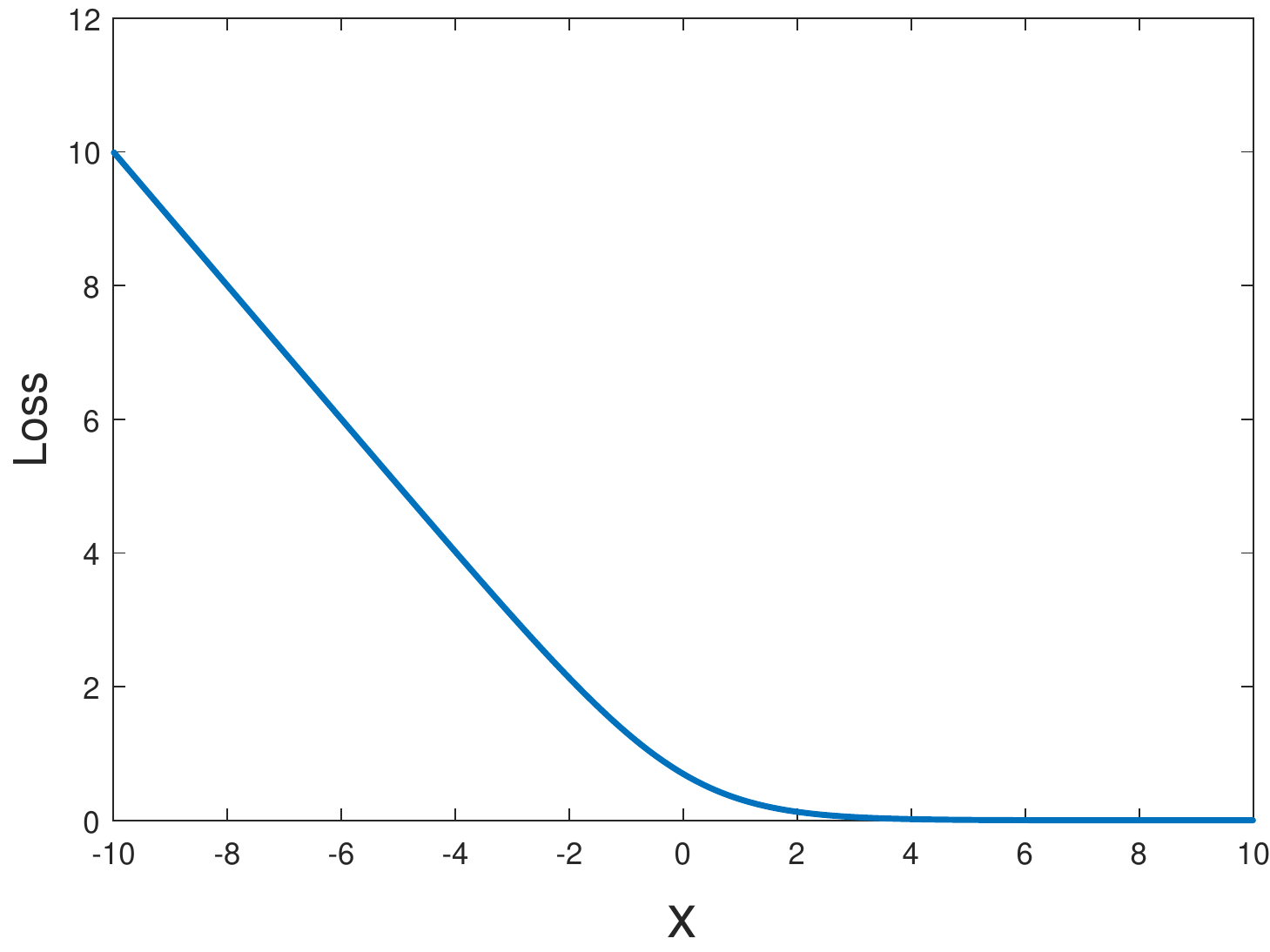}
&
 \includegraphics[width=1.5in]{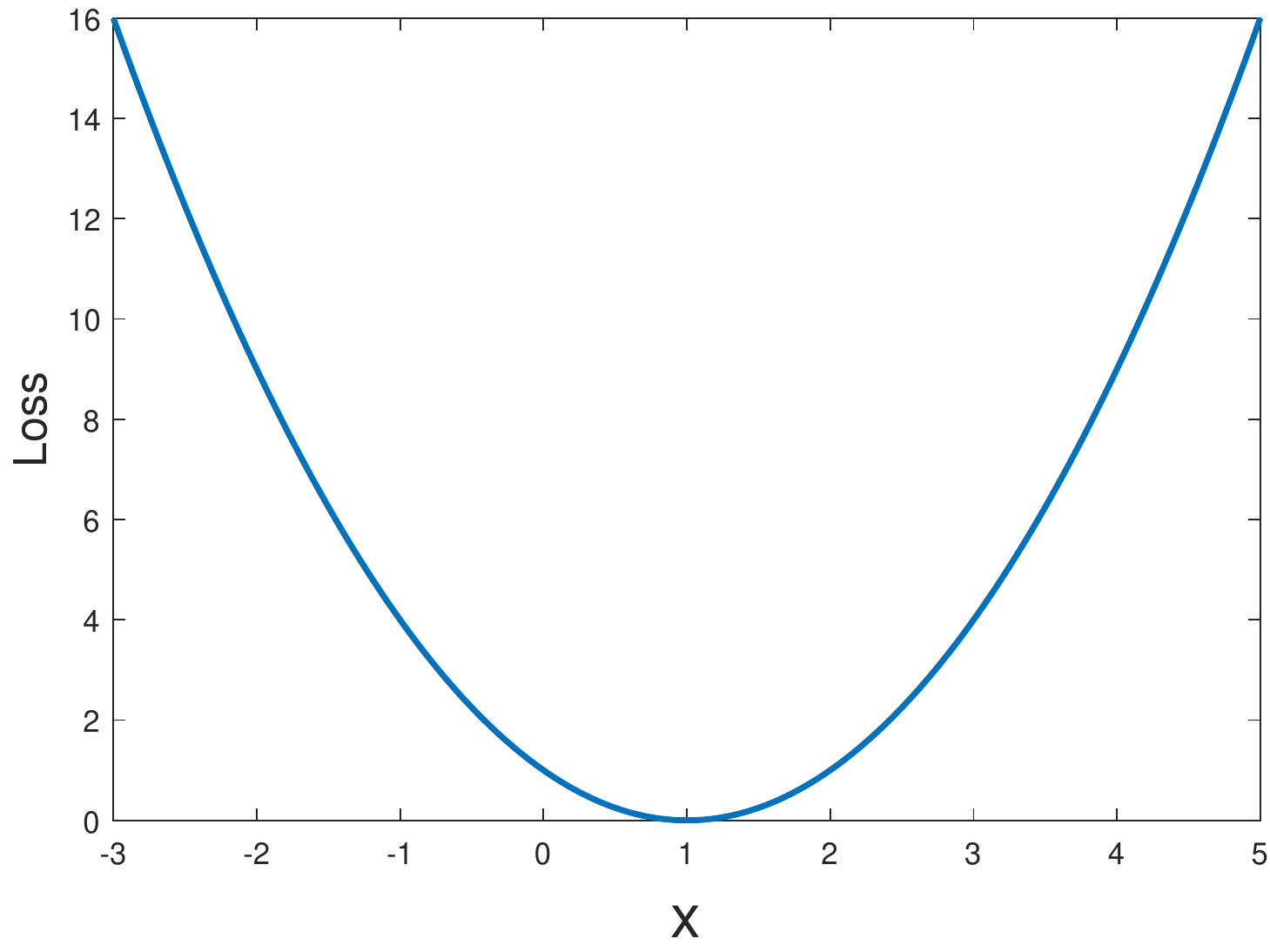}
\\
(a)
&
(b)
\end{tabular}
\caption{
(a): The sigmoid cross entropy loss function. (b): The least squares loss function.
}
\label{fig:loss}
\end{figure}

\subsection{Generative Adversarial Networks} 
The learning process of the GANs is to train a discriminator $D$ and a generator $G$ simultaneously. The target of $G$ is to learn the distribution $p_g$ over data $\bm{x}$. $G$ starts from sampling input variables $\bm{z}$ from a uniform or Gaussian distribution $p_z(\bm{z})$, then maps the input variables $\bm{z}$ to data space $G(\bm{z}; \theta_g)$ through a differentiable network. On the other hand, $D$ is a classifier $D(\bm{x}; \theta_d)$ that aims to recognize whether an image is from training data or from $G$. The minimax objective for GANs can be formulated as follows:

\begin{equation}
\label{eq_gan}
\begin{split}
\min_G \max_D V_{\text{\tiny GAN}}(D, G) = \mathbb{E}_{\bm{x} \sim p_{\text{data}}(\bm{x})}[\log D(\bm{x})] + \mathbb{E}_{\bm{z} \sim p_{\bm{z}}(\bm{z})}[\log (1 - D(G(\bm{z})))]&.
\end{split}
\end{equation}

\subsection{Least Squares Generative Adversarial Networks}
\label{sec:lsgan}
Viewing the discriminator as a classifier, regular GANs adopt the sigmoid cross entropy loss function. As stated in Section \ref{sec:introduction}, when updating the generator, this loss function will cause the problem of vanishing gradients for the samples that are on the correct side of the decision boundary, but are still far from the real data. To remedy this problem, we propose the Least Squares Generative Adversarial Networks (LSGANs). Suppose we use the $a$-$b$ coding scheme for the discriminator, where $a$ and $b$ are the labels for fake data and real data,  respectively. Then the objective functions for LSGANs can be defined as follows:

\begin{equation}
\label{eq:lsgan}
\begin{split}
\min_D V_{\text{\tiny LSGAN}}(D) = &\frac{1}{2}\mathbb{E}_{\bm{x} \sim p_{\text{data}}(\bm{x})}\bigl[(D(\bm{x})-b)^2\bigr]+ \frac{1}{2}\mathbb{E}_{\bm{z} \sim p_{\bm{z}}(\bm{z})}\bigl[(D(G(\bm{z}))-a)^2\bigr] \\
\min_G V_{\text{\tiny LSGAN}}(G) = &\frac{1}{2}\mathbb{E}_{\bm{z} \sim p_{\bm{z}}(\bm{z})}\bigl[(D(G(\bm{z}))-c)^2\bigr],
\end{split}
\end{equation}
where $c$ denotes the value that $G$ wants $D$ to believe for fake data.

\subsubsection{Benefits of LSGANs}
The benefits of LSGANs can be derived from two aspects. First, unlike regular GANs, which cause almost no loss for samples that lie in a long way on the correct side of the decision boundary (Figure \ref{fig:boundary}(b)), LSGANs will penalize those samples even though they are correctly classified (Figure \ref{fig:boundary}(c)). When we update the generator, the parameters of the discriminator are fixed, i.e., the decision boundary is fixed. As a result, the penalization will make the generator to generate samples toward the decision boundary. On the other hand, the decision boundary should go across the manifold of real data for a successful GANs learning. Otherwise, the learning process will be saturated. Thus moving the generated samples toward the decision boundary leads to making them be closer to the manifold of real data. 

Second, penalizing the samples lying a long way to the decision boundary can generate more gradients when updating the generator, which in turn relieves the problem of vanishing gradients. This allows LSGANs to perform more stable during the learning process. This benefit can also be derived from another perspective: as shown in Figure \ref{fig:loss}, the least squares loss function is flat only at one point, while the sigmoid cross entropy loss function will saturate when $x$ is relatively large.

\subsubsection{Reltation to f-divergence}
In the original GAN paper~\cite{Goodfellow2014}, the authors has shown that minimizing Equation \ref{eq_gan} yields minimizing the Jensen-Shannon divergence:
\begin{equation}
\label{eq:gan_js}
\begin{split}
C(G) &=KL \left(p_\text{data} \left \| \frac{p_\text{data} + p_g}{2} \right. \right) + KL \left(p_g \left \| \frac{p_\text{data} + p_g}{2} \right. \right)-\log(4). 
\end{split}
\end{equation}
\normalsize
Here we also explore the relation between LSGANs and f-divergence. Consider the following extension of Equation \ref{eq:lsgan}:
\begin{equation}
\label{eq:general_lsgan}
\begin{split}
\min_D V_{\text{\tiny LSGAN}}(D) = &\frac{1}{2}\mathbb{E}_{\bm{x} \sim p_{\text{data}}(\bm{x})}\bigl[(D(\bm{x})-b)^2\bigr] + \frac{1}{2}\mathbb{E}_{\bm{z} \sim p_{\bm{z}}(\bm{z})}\bigl[(D(G(\bm{z}))-a)^2\bigr] \\
\min_G V_{\text{\tiny LSGAN}}(G) = &\frac{1}{2}\mathbb{E}_{\bm{x} \sim p_{\text{data}}(\bm{x})}\bigl[(D(\bm{x})-c)^2\bigr] + \frac{1}{2}\mathbb{E}_{\bm{z} \sim p_{\bm{z}}(\bm{z})}\bigl[(D(G(\bm{z}))-c)^2\bigr].
\end{split}
\end{equation}
Note that adding the term $\mathbb{E}_{\bm{x} \sim p_{\text{data}}(\bm{x})}[(D(\bm{x})-c)^2]$ to $V_{\text{\tiny LSGAN}}(G)$ does not change the optimal values since this term does not contain parameters of $G$.

We first derive the optimal discriminator $D$ for a fixed $G$ as below :
\begin{equation}
\label{eq:optimal_d}
D^*(\bm{x}) = \frac{bp_\text{data}(\bm{x})+ap_g(\bm{x})}{p_\text{data}(\bm{x})+p_g(\bm{x})}.
\end{equation}
In the following equations we use $p_\text{d}$ to denote $p_\text{data}$ for simplicity. Then we can reformulate Equation \ref{eq:general_lsgan} as follows:
\small
\begin{equation}
\label{eq:lsgan_divergence}
\begin{split}
&2C(G) = \mathbb{E}_{\bm{x} \sim p_{\text{d}}}\bigl[(D^*(\bm{x})-c)^2\bigr]+\mathbb{E}_{\bm{x} \sim p_{g}}\bigl[(D^*(\bm{x})-c)^2\bigr] \\
&=\mathbb{E}_{\bm{x} \sim p_{\text{d}}}
\left[
\bigl(\frac{bp_\text{d}(\bm{x})+ap_g(\bm{x})}{p_\text{d}(\bm{x})+p_g(\bm{x})}-c\bigr)^2
\right] +\mathbb{E}_{\bm{x} \sim p_{g}}
\left[
\bigl(\frac{bp_\text{d}(\bm{x})+ap_g(\bm{x})}{p_\text{d}(\bm{x})+p_g(\bm{x})}-c\bigr)^2
\right] \\
&=\int_{\mathcal{X}}p_\text{d}(\bm{x}) \bigl(\frac{(b-c)p_\text{d}(\bm{x})+(a-c)p_g(\bm{x})}{p_\text{d}(\bm{x})+p_g(\bm{x})}\bigr)^2 \textrm{d}x + \int_{\mathcal{X}}p_g(\bm{x}) \bigl(\frac{(b-c)p_\text{d}(\bm{x})+(a-c)p_g(\bm{x})}{p_\text{d}(\bm{x})+p_g(\bm{x})}\bigr)^2 \textrm{d}x \\
&=\int_{\mathcal{X}} \frac{\bigl((b-c)p_\text{d}(\bm{x})+(a-c)p_g(\bm{x})\bigr)^2}{p_\text{d}(\bm{x})+p_g(\bm{x})} \textrm{d}x \\
&=\int_{\mathcal{X}} \frac{\bigl((b-c)(p_\text{d}(\bm{x})+p_g(\bm{x}))-(b-a)p_g(\bm{x})\bigr)^2}{p_\text{d}(\bm{x})+p_g(\bm{x})} \textrm{d}x.
\end{split}
\end{equation}
\normalsize 
If we set $b-c=1$ and $b-a=2$, then
\begin{equation}
\label{eq:lsgan_divergence}
\begin{split}
2C(G)&=\int_{\mathcal{X}} \frac{\bigl(2p_g(\bm{x})-(p_\text{d}(\bm{x})+p_g(\bm{x}))\bigr)^2}{p_\text{d}(\bm{x})+p_g(\bm{x})} \textrm{d}x \\
&=\chi^2_\text{Pearson}(p_\text{d}+p_g\|2p_g),
\end{split}
\end{equation}
where $\chi^2_\text{Pearson}$ is the Pearson $\chi^2$ divergence. Thus minimizing Equation \ref{eq:general_lsgan} yields minimizing the Pearson $\chi^2$ divergence between $p_\text{d}+p_g$ and $2p_g$ if $a$, $b$, and $c$ satisfy the condtions of $b-c=1$ and $b-a=2$. 

\subsubsection{Parameters Selection}
One method to determine the values of $a$, $b$, and $c$ in Equation \ref{eq:lsgan} is to satisfy the conditions of $b-c=1$ and $b-a=2$, such that minimizing Equation \ref{eq:lsgan} yields minimizing the Pearson $\chi^2$ divergence between $p_\text{d}+p_g$ and $2p_g$. For example, by setting $a=-1$, $b=1$, and $c=0$, we get the following objective functions:
\begin{equation}
\label{eq:lsgan_peason}
\begin{split}
\min_D V_{\text{\tiny LSGAN}}(D) = &\frac{1}{2}\mathbb{E}_{\bm{x} \sim p_{\text{data}}(\bm{x})}\bigl[(D(\bm{x})-1)^2\bigr] + \frac{1}{2}\mathbb{E}_{\bm{z} \sim p_{\bm{z}}(\bm{z})}\bigl[(D(G(\bm{z}))+1)^2\bigr] \\
\min_G V_{\text{\tiny LSGAN}}(G) = &\frac{1}{2}\mathbb{E}_{\bm{z} \sim p_{\bm{z}}(\bm{z})}\bigl[(D(G(\bm{z})))^2\bigr].
\end{split}
\end{equation}

Another method is to make $G$ generate samples as real as possible by setting $c=b$. For example, by using the $0$-$1$ binary coding scheme, we get the following objective functions:
\begin{equation}
\label{eq:lsgan_01}
\begin{split}
\min_D V_{\text{\tiny LSGAN}}(D) = &\frac{1}{2}\mathbb{E}_{\bm{x} \sim p_{\text{data}}(\bm{x})}\bigl[(D(\bm{x})-1)^2\bigr] + \frac{1}{2}\mathbb{E}_{\bm{z} \sim p_{\bm{z}}(\bm{z})}\bigl[(D(G(\bm{z})))^2\bigr] \\
\min_G V_{\text{\tiny LSGAN}}(G) = &\frac{1}{2}\mathbb{E}_{\bm{z} \sim p_{\bm{z}}(\bm{z})}\bigl[(D(G(\bm{z}))-1)^2\bigr].
\end{split}
\end{equation}

In practice, we observe that Equation \ref{eq:lsgan_peason} and Equation \ref{eq:lsgan_01} show similar performance. Thus either one can be selected. In the following sections, we use Equation \ref{eq:lsgan_01} to train the models.

\begin{figure}[t]
\centering
\begin{tabular}{c@{\hspace{0.6in}}c}
 \includegraphics[width=1.1in]{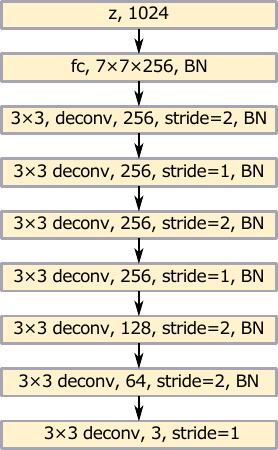}
&
 \includegraphics[width=1.1in]{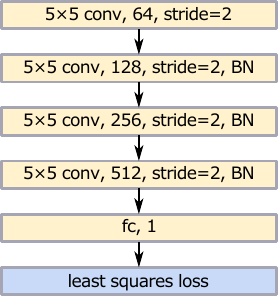}
\\
(a)
&
(b)
\end{tabular}
\caption{
 Model architecture. ``$K \times K$, conv/deconv, $C$, stride = $S$'' denotes a convolutional/deconvolutional layer with $K \times K$ kernel, $C$ output filters and stride = $S$. The layer with BN means that the layer is followed by a batch normalization layer. ``fc, $N$'' denotes a fully-connected layer with $N$ output nodes. The activation layers are omitted. (a):  The generator. (b): The discriminator.
}
\label{fig:arch1}
\end{figure}

\begin{figure}[h]
\centering
\begin{tabular}{cc}
 \includegraphics[width=2in]{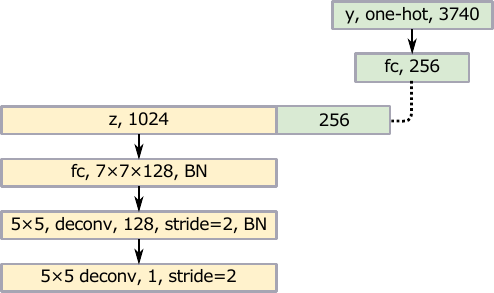}
 &
 \includegraphics[width=2in]{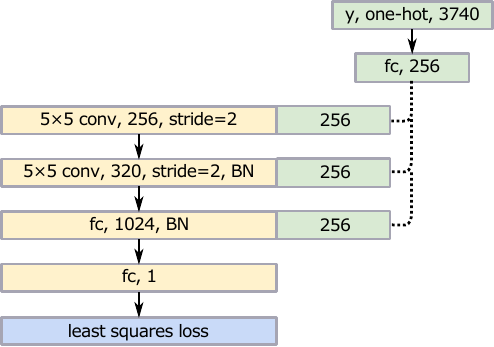}
\\
(a)
&
(b)
\end{tabular}
\caption{
 Model architecture for datasets with amount of classes. (a): The generator. (b): The discriminator.
}
\label{fig:arch2}
\end{figure}

\subsection{Model Architectures}
\label{sec:arch}
The first model we have designed is shown in Figure \ref{fig:arch1}, which is motivated by the VGG model~\cite{Simonyan2014}. Compared with the architecture in ~\cite{Radford2015}, two stride=$1$ deconvolutional layers are added after the top two deconvolutional layers. The architecture of the discriminator is identical to the one in~\cite{Radford2015} except for the usage of the least squares loss function. Following DCGANs, ReLU activations and LeakyReLU activations are used for the generator and the discriminator, respectively.

The second model we have designed is for tasks with lots of classes, for example, Chinese characters. For Chinese characters, we find that training GANs on multiple classes is not able to generate readable characters. The reason is that there are multiple classes in the input, but only one class in the output. As stated in ~\cite{Hornik1989}, there should be a deterministic relationship between input and output. One way to solve this problem is to use the conditional GANs~\cite{Mirza2014} because conditioning on the label information creates the deterministic relationship between input and output. However, using one-hot encoding, conditioning on the label vectors with thousands of classes is infeasible in terms of memory cost and computational time cost. We propose to use a linear mapping layer to map the large label vectors into small vectors first, and then concatenate the small vectors to the layers of models. In summary, the model architecture is shown in Figure \ref{fig:arch2}, and the layers to be concatenated are determined empirically. For this conditional LSGAN, the objective functions can be defined as follows:

\begin{equation}
\label{eq_clsgan}
\begin{split}
\min_D V_{\text{\tiny LSGAN}}(D) = &\frac{1}{2}\mathbb{E}_{\bm{x} \sim p_{\text{data}}(\bm{x})}\bigl[(D(\bm{x}|\Phi(\bm{y}))-1)^2\bigr] + \frac{1}{2}\mathbb{E}_{\bm{z} \sim p_{\bm{z}}(\bm{z})}\bigl[(D(G(\bm{z})|\Phi(\bm{y})))^2\bigr]. \\
\min_G V_{\text{\tiny LSGAN}}(G) = &\frac{1}{2}\mathbb{E}_{\bm{z} \sim p_{\bm{z}}(\bm{z})}\bigl[(D(G(\bm{z})|\Phi(\bm{y}))-1)^2\bigr],
\end{split}
\end{equation}
where $\Phi(\cdot)$ denotes the linear mapping function and $\bm{y}$ denotes the label vectors.

\begin{figure*}[t]
\centering
\begin{tabular}{c}
 \includegraphics[width=0.9\textwidth]{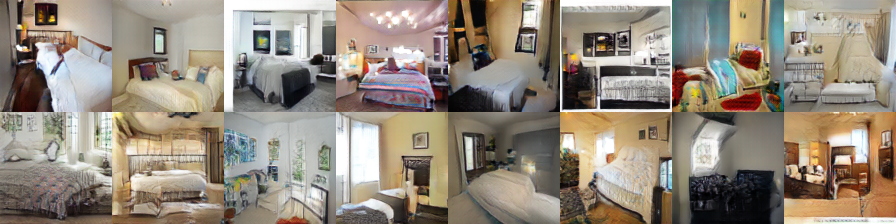}
\\
(a) Generated by LSGANs.
\\
 \includegraphics[width=0.9\textwidth]{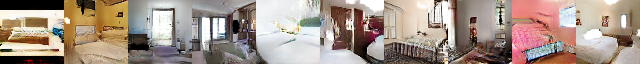}
\\
(b) Generated by DCGANs (Reported in~\cite{Radford2015}).
\\
 \includegraphics[width=0.9\textwidth]{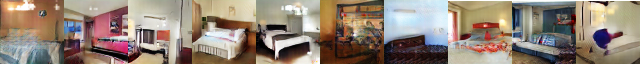}
\\
(c) Generated by EBGANs (Reported in~\cite{Zhao2016}).
\end{tabular}
\caption{
Generated images on LSUN-bedroom.
}
\label{fig:bedroom_cmp}
\end{figure*}

\section{Experiments}
\label{sec:experiments}

In this section, we first present the details of datasets and implementation. Next, we present the results of evaluating LSGANs on several scene datasets. Then we compare the stability between LSGANs and regular GANs by two comparison experiments. Finally, we evaluate LSGANs on a handwritten Chinese character dataset which contains $3740$ classes.

\begin{table}[h]
\small
\centering
\caption{Statistics of the datasets.}
\begin{tabular}{ccc}
\hline
Dataset & \#Samples & \#Categories\\
\hline
LSUN Bedroom&$3,033,042$&$1$\\
LSUN Church&$126,227$&$1$\\
LSUN Dining&$657,571$&$1$\\
LSUN Kitchen&$2,212,277$&$1$\\
LSUN Conference&$229,069$&$1$\\
HWDB1.0&1,246,991&3,740\\
\hline
\end{tabular}
\label{dataset}
\end{table}
\subsection{Datasets and Implementation Details}
We evaluate LSGANs on two datasets, LSUN~\cite{Yu2015} and HWDB1.0~\cite{Liu2011}. The details of the two datasets are presented in Table \ref{dataset}. The implementation of our proposed models is based on a public implementation of DCGANs\footnote{https://github.com/carpedm20/DCGAN-tensorflow} using TensorFlow \cite{tensorflow2015}. The learning rates for scenes and Chinese characters are set to $0.001$ and $0.0002$, respectively. Following DCGANs, $\beta_1$ for Adam optimizer is set to 0.5. All the codes of our implementation will be public available soon.

\begin{figure*}[t]
\centering
\begin{tabular}{cc}
 \includegraphics[width=0.45\textwidth]{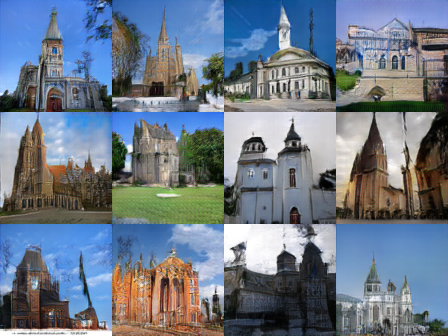}
&
 \includegraphics[width=0.45\textwidth]{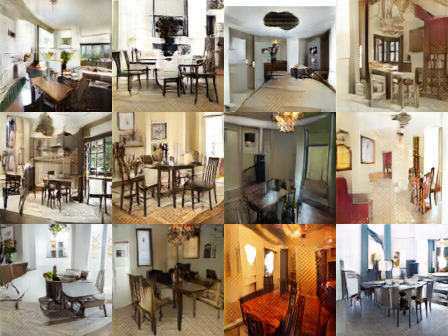}
\\
(a) \small Church outdoor.
&
(b) \small Dining room.
\\
 \includegraphics[width=0.45\textwidth]{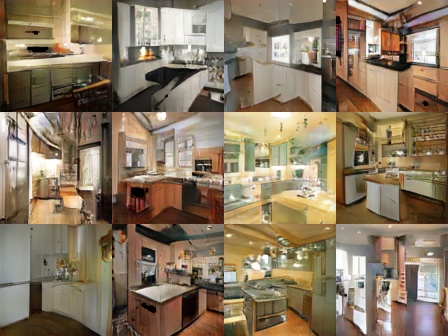}
&
 \includegraphics[width=0.45\textwidth]{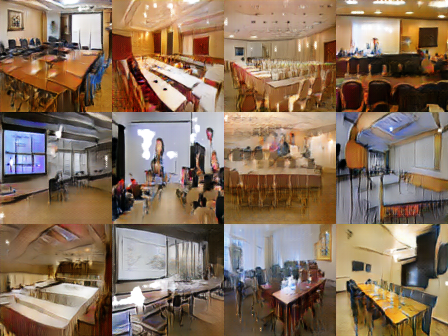}
\\
(c) \small Kitchen.
&
(d) \small Conference room.
\end{tabular}
\caption{
Generated images on different scene datasets.
}
\label{fig:scene}
\end{figure*}

\subsection{Scenes}
We train LSGANs (Figure \ref{fig:arch1}) on five scene datasets of LSUN including bedroom, kitchen, church, dining room and conference room. The bedroom generations by LSGANs and two baseline methods, DCGANs and EBGANs, are presented in Figure \ref{fig:bedroom_cmp}. We can observe that the images generated by LSGANs are of better quality than the ones generated by the two baseline methods. The results of LSGANs trained on other scene datasets are shown in Figure \ref{fig:scene}. 

\begin{figure*}[t]
\centering
\begin{tabular}{cc}
 \includegraphics[width=0.45\textwidth]{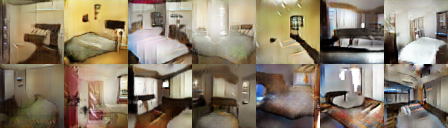}
&
 \includegraphics[width=0.45\textwidth]{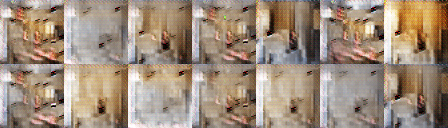}
\\
(a) \small LSGANs.
&
(b) \small Regular GANs.
\\
 \includegraphics[width=0.45\textwidth]{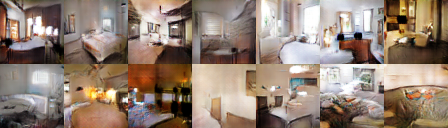}
&
 \includegraphics[width=0.45\textwidth]{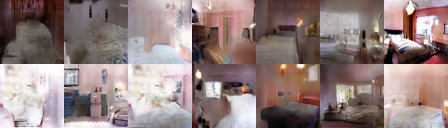}
\\
(c) \small LSGANs.
&
(d) \small Regular GANs.
\end{tabular}
\caption{
Comparison experiments by excluding batch normalization (BN). (a): LSGANs without BN in $G$ using Adam. (b): Regular GANs without BN in $G$ using Adam. (c): LSGANs without BN in $G$ and $D$ using RMSProp. (d): Regular GANs without BN in $G$ and $D$ using RMSProp.
}
\label{fig:no_BN}
\end{figure*}

\subsection{Stability Comparison}
As stated in Section \ref{sec:lsgan}, one benefit of LSGANs is the improved stability. Here we present two comparison experiments to compare the stability between LSGANs and regular GANs. 

One is to follow the comparison method in~\cite{Arjovsky2017}. Based on the network architectures of DCGANs, two architectures are designed to compare the stability. The first one is to exclude the batch normalization for the generator ($\text{BN}_G$ for short), and the second one is to exclude the batch normalization for both the generator and the discriminator ($\text{BN}_{GD}$ for short). As pointed out in ~\cite{Arjovsky2017}, the selection of optimizer is critical to the model performance. Thus we evaluate the two architectures with two optimizers, Adam~\cite{Kingma2014} and RMSProp~\cite{Tieleman2012}. In summary, we have four training settings, $\text{BN}_G$ with Adam, $\text{BN}_G$ with RMSProp, $\text{BN}_{GD}$ with Adam, and $\text{BN}_{GD}$ with RMSProp. We train these models on LSUN bedroom dataset using regular GANs and LSGANs separately and have the following four major observations. First, for $\text{BN}_G$ with Adam, there is a chance for LSGANs to generate relatively good quality images. We test $10$ times, and $5$ of them succeeds to generate relatively good quality images. But for regular GANs, we never observe successful learning. Regular GANs suffer from a severe degree of mode collapse. The generated images by LSGANs and regular GANs are shown in Figure \ref{fig:no_BN}. Second, for $\text{BN}_{GD}$ with RMSProp, as Figure \ref{fig:no_BN} shows, LSGANs generate higher quality images than regular GANs which have a slight degree of mode collapse. Third, LSGANs and regular GANs have similar performances for $\text{BN}_G$ with RMSProp and $\text{BN}_{GD}$ with Adam. Specifically, for $\text{BN}_G$ with RMSProp, both LSGANs and regular GANs learn the data distribution successfully, and for $\text{BN}_{GD}$ with Adam, both the ones have a slight degree of mode collapse. Last, RMSProp performs more stable than Adam since regular GANs are able to learn the data distribution for $\text{BN}_G$ with RMSProp, but fail to learn with Adam.

\begin{figure*}[t]
\centering
 \includegraphics[width=0.9\textwidth]{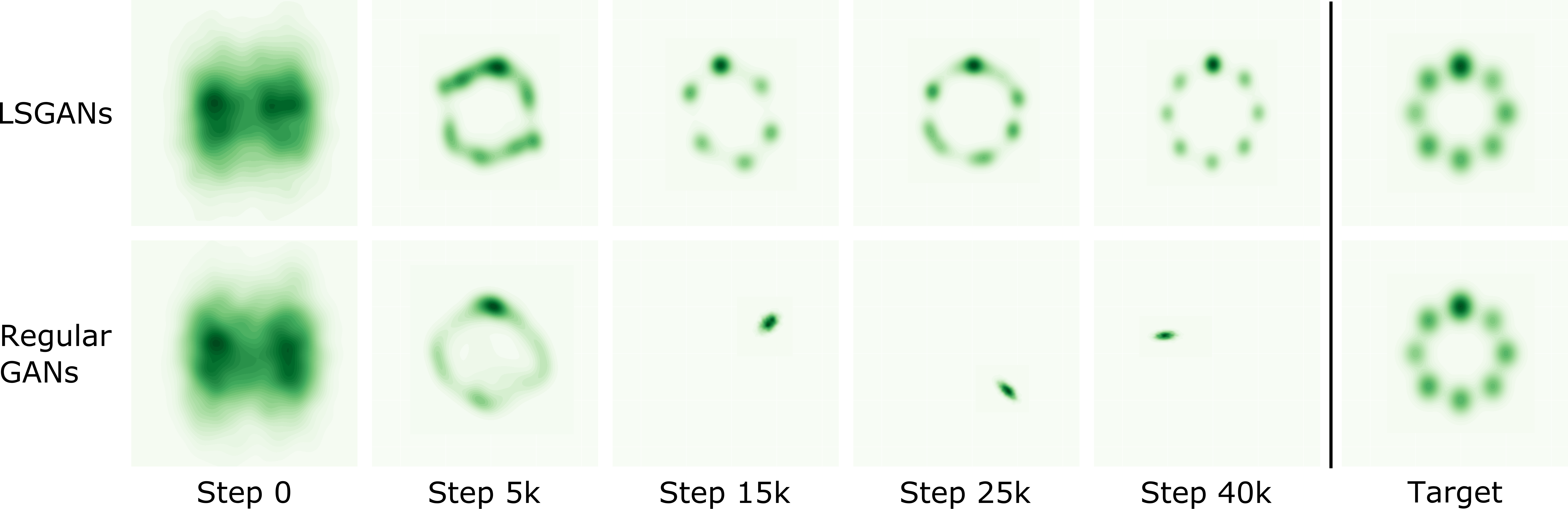}
\caption{
 Dynamic results of Gaussian kernel estimation for LSGANs and regular GANs. The final column shows the real data distribution. 
}
\label{fig:gaussian}

\end{figure*}

Another experiment is to evaluate on a Gaussian mixture distribution dataset, which is designed in literature ~\cite{Metz2016}. We train LSGANs and regular GANs on a 2D mixture of $8$ Gaussian dataset using a simple architecture, where both the generator and the discriminator contain three fully-connected layers. Figure \ref{fig:gaussian} shows the dynamic results of Gaussian kernel density estimation. We can see that regular GANs suffer from mode collapse starting at step $15$k. They generate samples around a single valid mode of the data distribution. But LSGANs learn the Gaussian mixture distribution successfully. 


\subsection{Handwritten Chinese Characters}
\begin{figure*}[t]
\centering
 \includegraphics[width=0.9\textwidth]{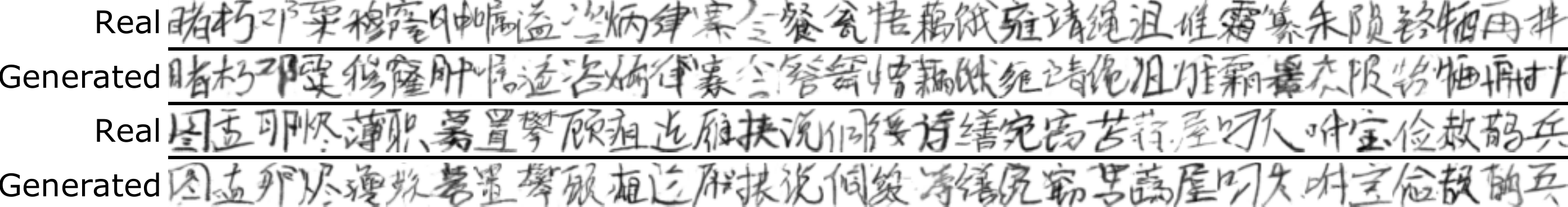}
\caption{
Generated images of handwritten Chinese characters by LSGANs. For row $1$ and row $2$, the images in the same column belong to the same class of characters. Row $3$ and row $4$ are also with this condition. The generated characters are readable.
}
\label{fig:chn_res}
\end{figure*}

We also train a conditional LSGAN model (Figure \ref{fig:arch2}) on a handwritten Chinese character dataset which contains $3740$ classes. LSGANs learn to generate readable Chinese characters successfully, and some randomly selected characters are shown in Figure \ref{fig:chn_res}. We have two major observations from Figure \ref{fig:chn_res}. First, the generated characters by LSGANs are readable. Second, we can get the correct labels of the generated images through label vectors, which can be used for further applications such as data augmentation. 


\section{Conclusions and Future Work}
\label{sec:conclusion}
In this paper, we have proposed the Least Squares Generative Adversarial Networks (LSGANs). Two model architectures (Figure \ref{fig:arch1} and Figure \ref{fig:arch2}) are designed. The first one is evaluated on several scene datasets. The experimental results show that LSGANs generate higher quality images than regular GANs. The second one is evaluated on a handwritten Chinese character dataset with $3740$ classes. Besides, numerous comparison experiments for evaluating the stability are conducted and the results demonstrate that LSGANs perform more stable than regular GANs during the learning process. Based on the present findings, we hope to extend LSGANs to more complex datasets such as ImageNet in the future. Instead of pulling the generated samples toward the decision boundary, designing a method to pull the generated samples toward the real data directly is also worth our further investigation.

\bibliography{lsgan}
\bibliographystyle{ieeetr}

\end{document}